\pdfoutput=1

\documentclass[11pt]{article}

\usepackage[final]{EMNLP2023}

\usepackage{times}
\usepackage{latexsym}
\usepackage{graphicx}
\usepackage{tabularx}
\usepackage[T1]{fontenc}
\usepackage{adjustbox}
\usepackage{listings}

\usepackage[utf8]{inputenc}

\usepackage{microtype}
\usepackage{graphicx}
\usepackage{amsmath}
\usepackage{inconsolata}
\usepackage{enumitem}
\usepackage{amsfonts}
\usepackage{amsmath}
\usepackage{soul}

\usepackage{comment}

\title{HR-MultiWOZ: A Task Oriented Dialogue (TOD) Dataset for HR LLM Agent}

\author{{\bf Weijie Xu}$^{1}$, {\bf Zicheng Huang}$^{1}$, {\bf Wenxiang Hu}$^{1}$, {\bf Xi Fang}$^{1}$, {\bf Rajesh Kumar Cherukuri}$^{1}$, \\
        {\bf Naumaan Nayyar}$^{1}$, {\bf Lorenzo Malandri}$^{2}$, {\bf Srinivasan H. Sengamedu}$^{1}$ \\
  $^{1}$Amazon\\
  $^{2}$University of Milano-Bicocca \\
  {\tt weijiexu@amazon.com}
}

\begin{document}
\maketitle
\begin{abstract}
Recent advancements in Large Language Models (LLMs) have been reshaping Natural Language Processing (NLP) task in several domains. Their use in the field of Human Resources (HR) has still room for expansions and could be beneficial for several time consuming tasks. Examples such as time-off submissions, medical claims filing, and access requests are noteworthy, but they are by no means the sole instances. However the aforementioned developments must grapple with the pivotal challenge of constructing a high-quality training dataset. On one hand, most conversation datasets are solving problems for customers not employees. On the other hand, gathering conversations with HR could raise privacy concerns. To solve it, we introduce HR-Multiwoz, a fully-labeled dataset of 550 conversations spanning 10 HR domains.  Our work has the following contributions:
(1) It is the first labeled open-sourced conversation dataset in the HR domain for NLP research. (2) It provides a detailed recipe for the data generation procedure along with data analysis and human evaluations. The data generation pipeline is transferable and can be easily adapted for labeled conversation data generation in other domains. (3) The proposed data-collection pipeline is mostly based on LLMs with minimal human involvement for annotation, which is time and cost-efficient. 


\end{abstract}

\section{Introduction}
Recent advances in natural language processing (NLP) have been applied in a variety of tasks in the Human Resources (HR) domain ranging from skill extraction~\citep{zhang2022skillspan}, job understanding~\citep{decorte2021jobbert} to candidate sourcing~\citep{hemamou-coleman-2022-delivering}. However, numerous HR processes remain highly inefficient, such as requesting time off, scheduling meetings, submitting tickets for IT issues, or filing medical claims. In fact, the Asana Work Index report shows that knowledge workers spend 60 percent of their time on repetitive work.

LLM agent~\citep{gao2023large} uses LLMs as its central computational engine, allowing it to carry on conversations, do tasks, reason, and display a degree of autonomy. Similar to other domains~\citep{kalvakurthi2023hey, hsu2023helping}, creating an LLM agent to help with these tasks could save a significant amount of time for employees and improve job satisfaction. A good LLM agent should be able to understand the requirements of users~\citep{liu2023agentbench}. The ideal dataset to evaluate or train an HR LLM agent should contain conversations between a virtual assistant and employees, annotated with dialogue states. Dialogue states contain representations of a conversation's current context such as intentions and relevant information. 
\begin{figure*}
\includegraphics[width=15cm, height = 5.7cm]{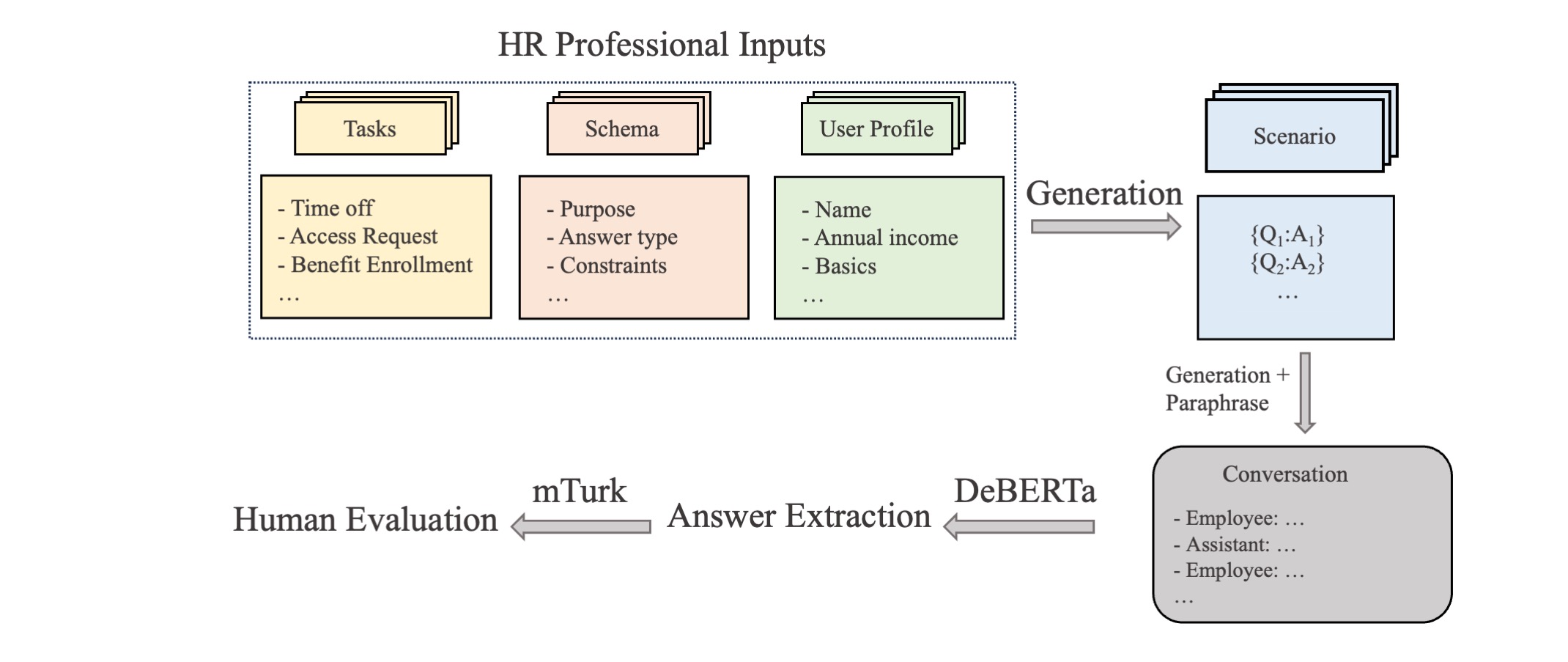}
\centering

\caption{The figure describes the data generation pipeline. The HR experts start by identifying tasks, creating schemas, and generating employee profiles. LLM is applied to generate diverse scenarios and paraphrase to make the conversation more natural. The label is then extracted by DeBERTa and refined by MTurk. }
\label{pipeline}
\end{figure*}

For a dataset to be useful in building/evaluating an HR LLM agent, it must satisfy the following four requirements:
(1) The information in the dialogue state must be \textbf{extractive}. When using an LLM agent to file a medical claim, employees must be able to trust that the system will accurately retrieve the right number. Thus, the extracted information must be from the conversation. 
(2) The information in the dialogue state should contain \textbf{long entity}. When using a LLM agent to solve a code bug, employees have to provide more detail about the code issue. This means that the extracted information should be long enough to give the LLM agent correct information. 
(3) The dataset must be \textbf{HR specific} and discuss about HR-relevant tasks.
(4) The conversation must be \textbf{empathetic}. In real conversations with HR, it is important to communicate with employees respectfully. This could enhance inclusive culture within the organization. The LLM agent built on this dataset could also be empathetic.

There are many open-source conversation datasets. Schema-Guided Dialogue (SGD) \citep{rastogi2020scalable} is a dialogue dataset with evolving ontologies, introducing new test set slots and services, emphasizing DST performance and zero-shot generalization. SGD-X \citep{Lee_2022} expands on SGD, presenting five additional schema styles. M2M \citep{shah2018building} connects a developer, who provides the task-specific information, and a framework, which provides the task-independent information, for generating dialogues
centered around completing the task. MultiWOZ~\citep{budzianowski2020multiwoz} features human-human dialogues using a stable ontology. However, all these datasets \textit{are customer-facing instead of employee-facing.} Also, \textit{none of them is fully extractive or related to HR}. \textit{The extracted information is also short in size.} HR LLM agent trained on these models may not be empathetic, extract complete information from employees and misunderstand employees' intent. Thus, it is essential to create a new dataset for HR application. On the other hand, collecting real datasets is difficult because the company cannot share these conversations with the public as it could leak employee confidential information. 
In response, we create an HR domain-specific dataset for LLM HR Agent. 
It is extractive, contains a long entity, is HR-specific, and contains empathetic conversations. We summarize our contributions as follows:\looseness-1
\begin{itemize}[leftmargin=*]
\item We've designed a data generation recipe that is efficient, cost-effective, high quality, and domain-specific. The same recipe can be easily adapted for labeled conversation generation in other domains.
\item We created a dataset of size 550 specifically for 10 HR use cases. The information in the dialogue state is extractive and contains a long entity.
\item The generated conversations are natural, clear, and empathetic based on human evaluations. The conversation is more comprehensive, detailed, richer in content, and diverse compared to existing datasets.
\end{itemize}

\section{Methods}
\label{sec:methods}
Our proposed data generation method is inspired by MultiWOZ~\citep{budzianowski2020multiwoz}. In Multiwoz, two annotators played the roles of a user and a wizard. The user was given a specific goal in a certain domain (like booking a hotel), and the wizard, having access to a database, responded to the user's requests. However, it requires a lot of human labeling which is expensive. With recent advancements of LLMs~\citep{brown2020language}, we can use LLMs to replace humans in generating more diverse scenarios and rephrasing conversations. At a high level, our generation process includes developing expert-validated HR schemas, generating diverse user profiles, creating realistic scenarios via Claude, randomizing and merging, rephrasing dialogues using Claude, and applying extractive modeling using DeBERTa model~\citep{he2021deberta} and human labeling to get high-quality labels. We chose Claude because of cost and ethical reasons as explained in Appendix~\ref{claude}. For each step, we provide detailed instructions and human labeling guidance in the Appendix. \textbf{This makes our data generation method easily transferable, reproducible and transparent.} It took 2 days and costed $38.32$ dollars for LLMs inference and 49.82 dollars for Human labeling. \textbf{This makes our method time and cost effective.} The detailed data generation pipeline is in Fig~\ref{pipeline} and a provided example is in Fig~\ref{example}.  

\textbf{Schema Creation} The input of the system includes diverse task schemas for different HR-related tasks. Each schema is composed of a series of structured questions, the schema's purpose, answer type, and constraints for each potential value. To ensure domain relevance and accuracy, these task schemas undergo a thorough audit by HR domain experts. The domain includes benefits enrollment, performance review, training requirements, safety incident report, relocation request, harassment report, goal setting, access request, it issues report and time off report. Each schema contains different slots. \textbf{This makes our generated dataset HR Specific.} For each slot, we also designed a question, answer type, and potential choices. The detailed example of task schema is in Table~\ref{tab:benefits}.

Next, we develop a user profile schema, focusing on the user's preference. This user profile schema aims to maximize diversity and represent a wide range of real-world scenarios. An example profile includes attributes such as Number of Dependents, Contact Preference, Annual Income, etc. These user profile schema were generated by Claude. We manually remove user profiles that share more than 2 entries with other profiles to maximize the diversity. A detailed example of a user profile is in Table~\ref{tab:person_info}. For company specific schema and user profile, company can adapt the same logic and modify the key and value to be company specific.

\textbf{Scenario Generation}
The scenario is the outline of the conversation. Taking the user profile and task schemas as input, we generate a realistic template as a Python dictionary (Question as key and generated answer from the selected user profile as value). We first complete the answer from the selected task schema using the user profile. Secondly, Claude is employed to answer the rest of the question in the scenario from the user's perspective. We instruct LLM to ensure that answers are concise yet informative. The detailed prompt is in Table~\ref{tab:user_template_instructions}.

\textbf{Conversation Generation and Paraphrasing}
To transform a scenario into a conversation, it should adopt a natural tone and structure. For instance, the conversation should be empathetic and includes expressions like "Cool", "Okay" etc. Also, in a real-world conversation, a user can sometimes answer multiple questions in one turn. For each template,  we then randomize the order of the scenario. We randomly combine answers of similar types into a single response. We then rewrite it as a question and answer. Finally, we use LLM to paraphrase questions and answers to enhance empathy in the questions and naturalness and completeness in responses. This paraphrase also provides a long entity such as a detailed description of a code error. Thus, \textbf{the paraphrased conversations are empathetic and the information in dialogue states contains long entity} The detailed prompt is in Table~\ref{tab:conversation_instructions}.

\textbf{Dialogue States Labeling}
The quality of the generated dialogues was assessed through answer extraction, data cleaning, and human evaluation. The answer was extracted using DeBERTa~\cite{he2021deberta} from the generated dialogues. This model is chosen for its compact size, effectiveness in extraction tasks, and capability to provide confidence scores between 0 and 1. We input questions, ground truth answers, and context into the model to extract answers with corresponding confidence levels. This step is crucial to ensure that answers in our dataset are not only informative but also extractable with a degree of certainty, which makes it easier to identify wrong answers. \textbf{This step makes the information in dialogue states extractive.}

The extracted answers were cleaned for use in the TOD system through a series of steps. We first remove all leading and trailing spaces, which often occurs as a byproduct of extraction processes. To align with the format of answers in the conversation template, we also remove all trailing punctuation marks. This step eliminates ambiguities and preserves the integrity. 

We further use mechanical turk to verify if the formatted extracted answer is equivalent to the answer in the scenario as asking a question in Figure~\ref{fig:mturk}. Following ~\citep{li-etal-2023-coannotating}, we selected the extracted answer with confidence below 0.1 for Mechanical Turk. This contains 692 data points. The answer can only be yes or no. We use 3 labelers per task and pay them 0.024 per task. If the response is 'no', we further label the data manually by HR professionals. Out of 71 data points marked as 'no', HR professionals identified 27 as inaccurately labeled and corrected them with the correct answer that is extractable from the conversation.

\section{Evaluation}

\textbf{Dataset Statistics:}
We are releasing the HR Multiwoz dataset comprising a total of $550$ dialogues collected using the proposed method in Sec. \ref{sec:methods}. This dataset covers HR-related tasks including benefits enrollment, performance review, training requirements, safety incident reports, relocation requests, harassment reports, goal settings, access requests, IT issues reporting, and time off reports as shown in Table. \ref{tab:dataset}. Our dataset covers diverse topics in HR and provides a wide range of examples. Thus, compared to the existing dataset, we recommend using this dataset for transfer learning tasks in other HR-related use cases.


\textbf{Datasets Comparison:}
Compared to the existed dataset, the HR Multiwoz dataset exhibits diversity and completeness in questions and answers, as illustrated in Table~\ref{tab:evaluation}. The dataset contains fewer dialogues than the M2M restaurant dataset, yet it surpasses it in total turns and total tokens. \textbf{This indicates that the HR Multiwoz dialogues are extended and richer in content.} HR Multiwoz achieves the highest average turns per dialogue and average tokens per turn. This suggests that \textbf{the conversations are both comprehensive and detailed.} 

Furthermore, the highest ratios of unique tokens and unique bigrams in our dataset signify a broader verbal crucial for natural responses. Such diversity in language use is indicative of the dataset’s capacity to simulate real-world conversations in the HR-specific domain. Additionally, \textbf{the inclusion of long entities in user answers, as suggested by the highest average tokens per answer}, enhances the dataset's utility for training sophisticated dialogue systems that require an understanding of extended contexts and nuanced language. Overall, the HR Multiwoz dataset appears to be well-suited for developing/evaluating HR LLM Agents that can effectively handle empathetic, natural, and complete interactions in HR-specific scenarios.


\begin{table}[ht!]
\centering
\begin{tabularx}{\linewidth}{|X|X|X|X|}
\hline
\textbf{Metrics} & \textbf{multiwoz} & \textbf{M2MR} & \textbf{ours} \\
\hline
Dialogues & 8437 & 1116 & 550 \\
\hline
Total turns & 113552 & 6188 & 8910 \\
\hline
Total tokens & 1742157 & 99932 & 181363 \\
\hline
Avg. turns per dialogue & 13.46 & 11.09 & \textbf{16.2} \\
\hline
Avg. tokens per turn & 15.34 & 8.07 & \textbf{20.35} \\
\hline
Avg. tokens per answer & 13.46 & 5.56 & \textbf{14.53} \\
\hline
Unique tokens / Total tokens & 0.0103 & 0.0092 & \textbf{0.0156} \\
\hline
Unique bigrams / Total tokens & 0.0634 & 0.0670 & \textbf{0.1177} \\
\hline
\end{tabularx}
\caption{Comparing Multiwoz 2.2,  M2M Restaurants and our datasets: HR-MultiWOZ on diversity of language and dialogue flows.}
\label{tab:evaluation}
\end{table}
\textbf{Human Evaluations:}
In the subjective evaluation of the Multiwoz dataset, crowd workers assessed the naturalness of employees' responses, the clarity of HR's questions, and the politeness of HR's questions. For each category, only responses with confidence scores above 60 percent were considered, resulting in 634 employee answers, 623 HR questions for clarity, and 629 HR questions for politeness in the evaluation set. Statistical analysis using one-sample t-tests revealed that the average ratings for employees' naturalness, HR's question clarity, and HR's question politeness were significantly higher than neutral. This was evidenced by high t-statistics (19.31 for naturalness, 18.83 for clarity, and 16.02 for politeness) and extremely low p-values, indicating \textbf{strong positive ratings in the naturalness of employees' responses, the clarity of HR's questions, and the politeness of HR's questions.} The detailed score distributions, instructions and detailed analysis are in Appendix~\ref{humaneval}, Appendix~\ref{humaninstruct} and Appendix~\ref{human}.

\section{Conclusions}
HR-Multiwoz, our generated dataset of 550 labeled conversations, can evaluate/train HR LLM agents by offering 10 domain-specific, diverse, comprehensive, detailed labeled conversations. Our data generation approach minimizes human annotation efforts while maximizing data relevance and quality, leveraging Claude. This makes our data generation approach transferable. As the first dataset in HR dialogue systems, HR-Multiwoz represents a significant advancement in HR automation, providing rich and empathetic dialogues ideal for training efficient, human-like HR digital assistants. It satisfies all HR dialogue requirements and sets a new benchmark for HR applications, paving the way for innovative, AI-driven HR solutions. In the future, we suggest to enhance this dataset by increasing the number of conversations, extending to other languages beyond English, and including suggest API to call at the end of each conversation. We will use cc-by-4.0 license. 
We provide ethical statement and limitations in Appendix~\ref{ethics} and Appendix~\ref{limit}

\bibliography{anthology,custom}
\bibliographystyle{acl_natbib}

\appendix
\section{Ethics Statement}
\label{ethics}
Ethics Statement The dataset generated by AI in the HR space necessitates careful consideration of ethical issues related to safety, privacy, and bias. There is a possibility that, in attempting to assist, AI generated dataset may cause more harm than benefit. In response, in collaboration with security reviewer and HR professionals, we have taken the following steps in order to minimize the risks of harm.

Human Labeling: To make sure the generated conversation is polite and empathetic, we use human labelers to label the conversation. 

Guardrail: We remove conversation that is labeled by human that contains rude language. This makes sure the language is not rude. 

Privacy: In our generated data, we use synthetic user profile which is not real. We also make sure the data in the system is in compliance with rigorous internal infoSec policies and standards. 

Negative Examples/Potential Bias: To mitigate potential biases in generative models, we have employed an extractive approach. None the less, the effectiveness of extractions could vary with the employee's language fluency. This variation could potentially lead to inefficiencies in the Task-Oriented Dialogue (TOD) system for non-native English speakers. Efforts are underway to understand and address these issues.

Synthetic Data Bias: The dataset primarily relies on conversations generated through large language models (LLMs) and human rephrasing. This may introduce biases inherent in the LLMs or limit the scenarios to the creative constraints of the model's training data.

Limited Cultural and Linguistic Diversity: HR-Multiwoz may primarily reflect the cultural and linguistic norms of the data creators or the LLM training data. This limitation could affect the dataset's effectiveness in global or culturally diverse HR settings.

\section{Limitations}
\label{limit}
Updating and expanding the dataset to include new HR domains or to adapt to evolving HR practices and policies might require some efforts, given the reliance on new schema creation.  The dataset does not contains task part of the conversation. This limits the use of this dataset to train an LLM agent to leverage different tools. This dataset also lacks evaluations on existed TOD systems method.

DeBERTa model also has some limitations. We observe additional complexities when comparing the original short answers with those extracted by DeBERTa, for example: (i) duplication of answers in a single turn containing multiple short answers, (ii) inclusion of prompting text like "Employee:", (iii) failure to extract meaningful answers or labels.

The performance of generated datasets is not fully controllable. human feedback is essential to further improve the dataset. With this regards, LLMs allow the user to be informed with the final outcome of the system (you have been assigned a time-off period from…to…) and check the correctness of the process.

\section{Future Work}
Real-World Integration and Testing: Implementing the model trained on this dataset in real-world HR environments to test and refine its efficacy. This could include pilot programs with HR departments to gather feedback and improve the dataset's realism and applicability.

Cross-Cultural and Multilingual Expansion: Enhancing the dataset to include a broader range of cultural contexts and languages, making it more inclusive and applicable globally, especially in diverse workplaces.

Continuous Updating and Expansion: Regularly updating the dataset to reflect the latest HR practices, policies, and regulations. This could involve creating a framework for continuous data collection and integration.

Bias Detection and Mitigation: Implementing systematic methods to identify and mitigate biases in the dataset, ensuring fair and unbiased HR-related dialogues.

Broader Domain Generalization: Extending the dataset or its methodology to other domains beyond HR, thereby testing its adaptability and utility in various fields like customer service, healthcare, or legal advice.

User Experience Research: Conducting user experience research to understand how employees and HR professionals interact with AI-based systems trained on the dataset, aiming to improve user satisfaction and effectiveness.

Topic Modeling: Leveraging topic modeling techniques to understand the theme in these conversations.~\citep{xu-etal-2023-vontss, xu-etal-2023-detime} 

Differentially Private Dataset: Make sure dataset is fair and privacy preserving.~\citep{Xu2021}

\section{Task Profile}
\begin{table*}
\centering
\label{prompt}
\begin{tabular}{|l|p{10cm}|}
\hline

\textbf{Key} & \textbf{Description} \\ \hline
type\_of\_benefit & What type of benefit do you want to enroll in? (e.g., Health Insurance, Dental Insurance, etc.) \\ \hline
benefit\_plan\_selection & Select your benefit plan by entering the plan code (e.g., Plan A, Plan B, etc.). \\ \hline
number\_of\_dependents & How many dependents do you want to add to the plan? (Enter a number) \\ \hline
previous\_coverage\_duration & How many years have you been previously covered under a health plan? (Enter a number) \\ \hline
effective\_date & When do you want the coverage to start? (Enter the date in YYYY-MM-DD format) \\ \hline
personal\_information\_confirmation & Do we have your updated personal information on file? (Answer with Yes or No) \\ \hline
contact\_preference & Please enter your preferred contact method (Email, Phone, Mail). \\ \hline
estimated\_annual\_premium & What is your estimated annual premium budget in USD? (Enter a number) \\ \hline

\end{tabular}
\caption{Benefits Enrollment Schema Example. This is just an example. Each question could involve multiple types.}
\label{tab:benefits}
\end{table*}

\begin{table*}[ht]
\centering
\begin{tabular}{|l|l|}
\hline
\textbf{Key}                  & \textbf{Value}                \\ \hline
Number of Dependents          & 2                             \\ \hline
Contact Preference            & Email                         \\ \hline
Annual Income                 & \$150,000                     \\ \hline
Name                          & Dr. Li Wei                    \\ \hline
Contact Information           & liwei@medicalemail.com        \\ \hline
Current Location              & San Francisco, CA             \\ \hline
Job                           & Doctor                        \\ \hline
\end{tabular}
\caption{User Profile Example}
\label{tab:person_info}
\end{table*}

\begin{table*}[ht]
\centering
\begin{tabular}{|p{10cm}|}
\hline
\textbf{Instruction}  \\ 
User: \{user\}  \\ 
Template: \{template\} \\
You are User. \\
Fill out all questions in template based on experience. \\
Generated dictionary should contain key name and generated answer. \\
All keys from Template are in generated dictionary. \\ 
Make the answer extremely short (within 5 words). \\ 
Put the generated dictionary in \textless answer\textgreater\textless /answer\textgreater XML tags.\\ \hline
\end{tabular}
\caption{Instructions for Template Generation}
\label{tab:user_template_instructions}
\end{table*}

\begin{table*}[ht]
\centering
\begin{tabular}{|p{10cm}|}
\hline
\textbf{Instruction}  \\ 
Conversation: \{conversation\}  \\ 
This is the conversation between HR Assistant and an employee. \\ 
1. For each Question, paraphrase the question to make it more conversational by using more modal words and empathetic. \\
2. For each Answer, write it as a complete sentence. \\
Please put the updated Conversation based on Template in \textless answer\textgreater\textless /answer\textgreater XML tags.\\ \hline
\end{tabular}
\caption{Instructions for Conversation Rewriting}
\label{tab:conversation_instructions}
\end{table*}

\section{Human Evaluation Analysis}
\label{human}
For a subjective evaluation of the Multiwoz dataset, we want to understand the following: 1. Is the employees' answer natural? 2. Is the HR's question clear? 3. Is the HR's question polite or empathetic?  We presented final dialogues to crowd workers, who rated each user and HR turn on a scale of 1 to 5 for the specific dimensions with 1 being very robotic and 5 being very natural. We sampled 650 turns from HR and employees to create the evaluation set. Each turn was shown to 3 crowd workers. We pay them 0.012 per task. Each answer also has a confidence score between 0 to 1, indicating the labelers' confidence in their assessment.

For question 1, we included data with confidence score larger than 60, resulting in 634 HR turns to create the evaluation set. Using a one-sample t-test, we showed that the average rating is significantly better than neutral indicating that the question from the employee's answer is natural (t-statistic around $19.31$, p value $\leq 0.000000001$.

For question 2, We only select confidence score larger than 60 which is 623 turn from HR to create the evaluation set. Score 3 is neutral. Score 5 is very clear. 1 is very unclear. The one-sample t-test to evaluate if the average is significantly better than neutral. The test gives a t-statistic of approximately $18.83$ and an extremely small p-value ( p value $\leq 0.000000001$). This result indicates that the average rating is significantly better than neutral indicating that the question from HR is clear.  

For question 3,  We only select confidence score larger than 60 which is 629 turn from HR to create the evaluation set. Score 3 is neutral. Score 5 is very polite. 1 is very rude. The one-sample t-test to evaluate if the average is significantly better than neutral. The test gives a t-statistic of approximately $16.02$ and an extremely small p-value ( p value $\leq 0.000000001$). This result indicates that the average rating is significantly better than neutral indicating that the question from HR is polite. 

\section{Example of Data Generation Process}
\begin{figure*}
\includegraphics[width=15cm]{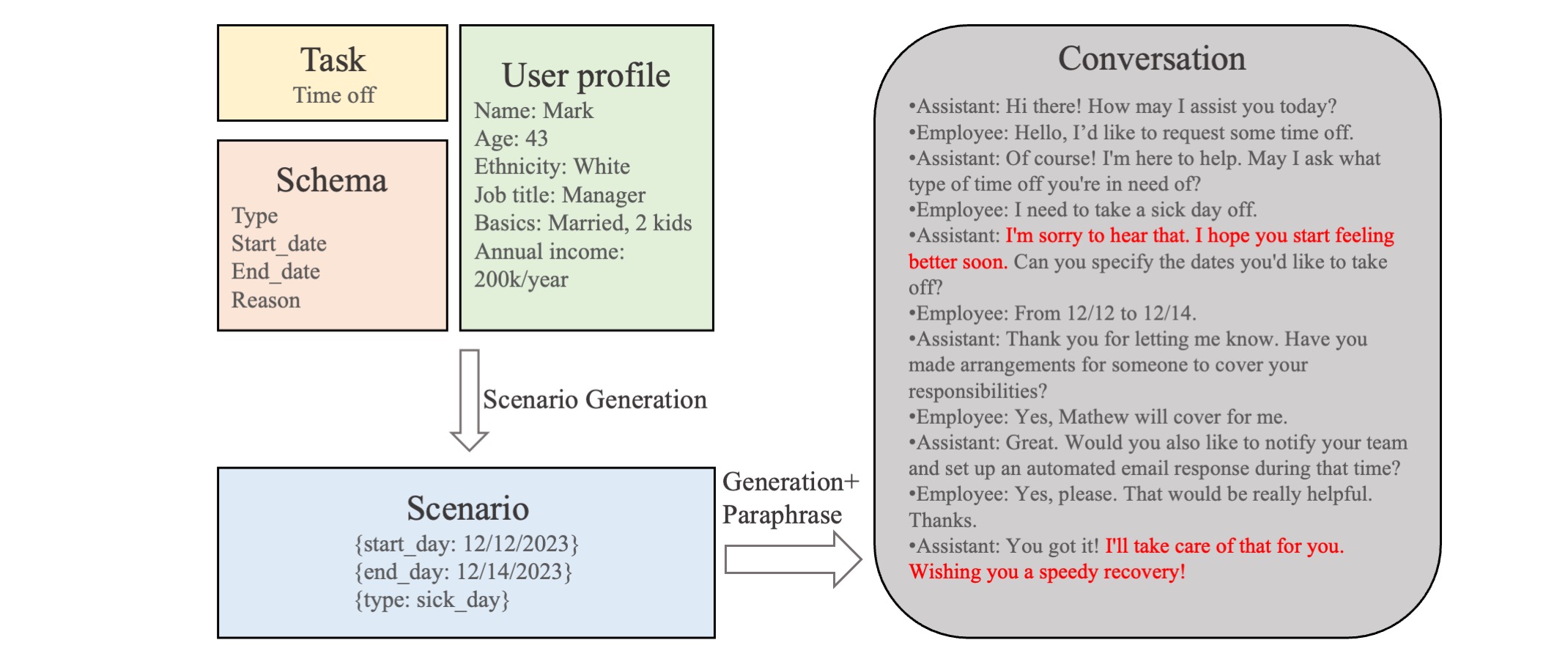}
\centering

\caption{The figure describes a conversation generation process. We first identify task, schema and employee profile. We then use LLM to fill out the value in the schema. We then use LLM to rephrase the conversation to be more natural. We highlight the part that HR assistant show empathy in red. }
\label{example}
\end{figure*}

\section{Claude}
\label{claude}
We choose claude over GPT4 for the following reasons:
\textbf{Cost Effiency}: Claude is more cost-effective in terms of computing resources required for data generation. For instance, using the GPT-4 8K context model via OpenAI's API costs \$0.03 for every 1K input tokens and \$0.06 for every 1K output tokens. 
\textbf{Data Privacy and Security} Claude offers better data privacy and security features, especially for sensitive tasks like generating data for HR-related applications. 
\textbf{Model Characteristics} Claude is trained RLAIF which could produce more ethical conversations.

\section{Generated Dataset Statistics}
\begin{table}[ht]
\centering
\begin{tabularx}{\linewidth}{|X|X|}
\hline
\textbf{Domain} & Number of dialogues\\
\hline
benefits\_enrollment & 71 \\
\hline
performance\_review & 64 \\
\hline
training\_request & 57 \\
\hline
safety\_incident\_report & 103 \\
\hline
relocation\_request & 57 \\
\hline
harassment\_report & 103 \\
\hline
goal\_setting & 42 \\
\hline
access\_request & 278 \\
\hline
it\_issue\_report & 38 \\
\hline
time\_off\_report & 71 \\
\hline
\end{tabularx}
\caption{HR Multiwoz dataset.}
\label{tab:dataset}
\end{table}

\section{Example of Generated Dialogues}
\begin{table*}[ht]
\centering
\begin{tabular}{|p{10cm}|}
\hline
\textbf{One Example of Generated Dialogues}  \\ 
\textit{HR Assistant}: Hi there, how can I help you out today?  \\
\textit{Employee}: I'm hoping you can help me set up some machine learning training for my team.  \\
\textit{HR Assistant}: No problem, tell me a bit about what kind of training you're looking for. What will you guys be learning about and do you need any special accommodations?  \\
\textit{Employee}: We're hoping to improve our machine learning research skills. But no special accommodations are needed.  \\
\textit{HR Assistant}: Great, how many people will be attending this training? And approximately how long do you think it should last?  \\
\textit{Employee}: Three people will be participating and we'd like it to run for two days.  \\
\textit{HR Assistant}: Okay, thanks. Do you have a budget in mind for this? It would also help to know what your goals are for your team after the training.  \\
\textit{Employee}: Our budget is \$5000. We want to come away with better research skills.  \\
\textit{HR Assistant}: No problem, when works best for your schedule? Also, do you have a preference for how the training is delivered, like online or in-person?  \\
\textit{Employee}: We're hoping for some online seminars on June 15th, 2023.  \\
\textit{HR Assistant}: Sounds good. One last thing - would you say your team's current skill level is beginner, intermediate or advanced? And can you provide the contact info for the person coordinating it all?  \\
\textit{Employee}: Their level is intermediate. The coordinator is Henry Kim and you can reach him at hkim@researchlab.com or 555-1234.  \\
\hline
\end{tabular}
\caption{Example of Generated Dialogues}
\label{tab:example_dialogue}
\end{table*}

\section{Answer Evaluation}
        
\begin{figure*}
\includegraphics[width=16cm]{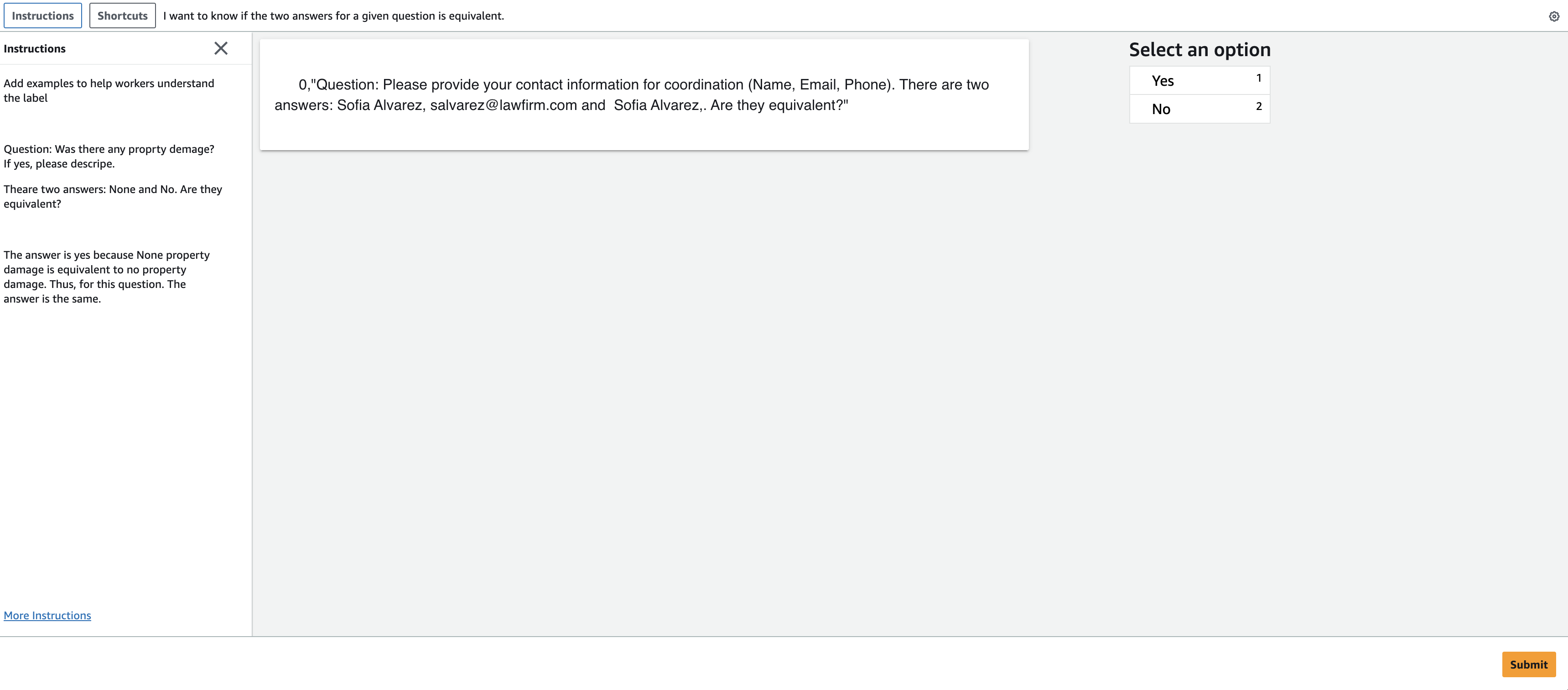}
\centering

\caption{MTurk Questions and selected examples to understand if extracted answer is equivalent to the ground truth}
\label{fig:mturk}
\end{figure*}

\section{Human Evaluation Score Distribution}
\label{humaneval}

\begin{figure*}
\includegraphics[width=16cm]{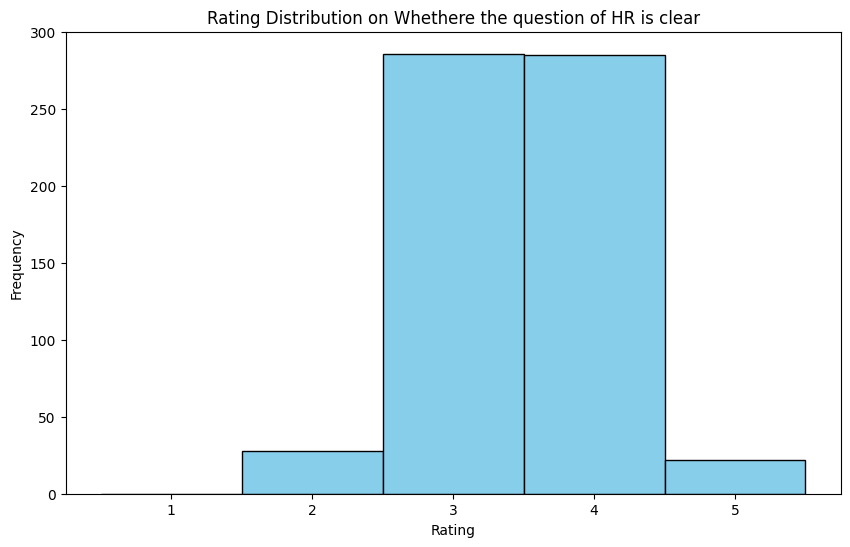}
\centering

\caption{MTurk Score Distribution to understand if the HR question is clear}
\end{figure*}

\begin{figure*}
\includegraphics[width=16cm]{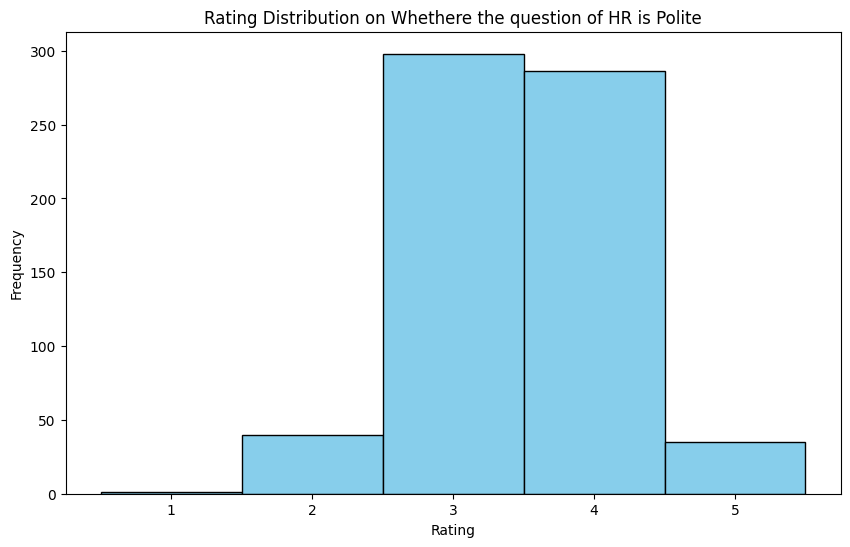}
\centering

\caption{MTurk Score Distribution to understand if the HR question is polite}
\end{figure*}

\begin{figure*}
\includegraphics[width=16cm]{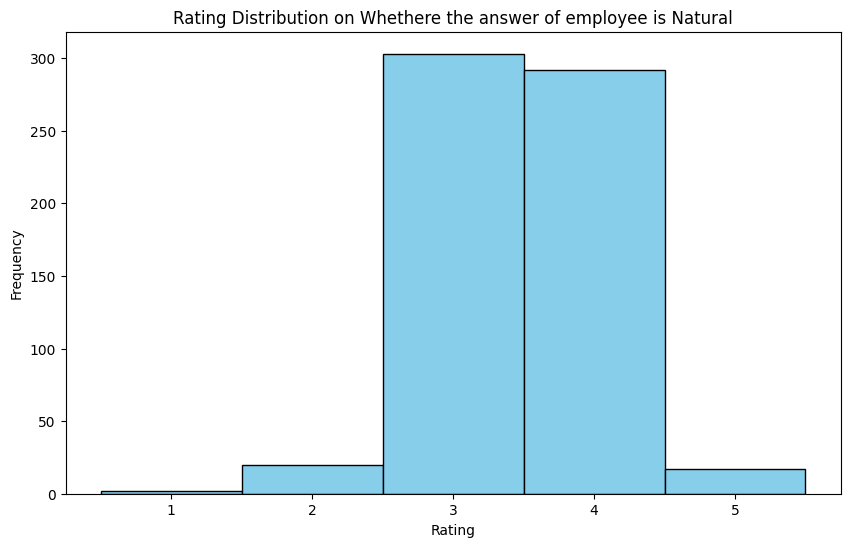}
\centering

\caption{MTurk Score Distribution to understand if the employee answer is natural}
\end{figure*}

\section{Human Evaluation Instructions }

\label{humaninstruct}

\begin{figure*}
\includegraphics[width=16cm]{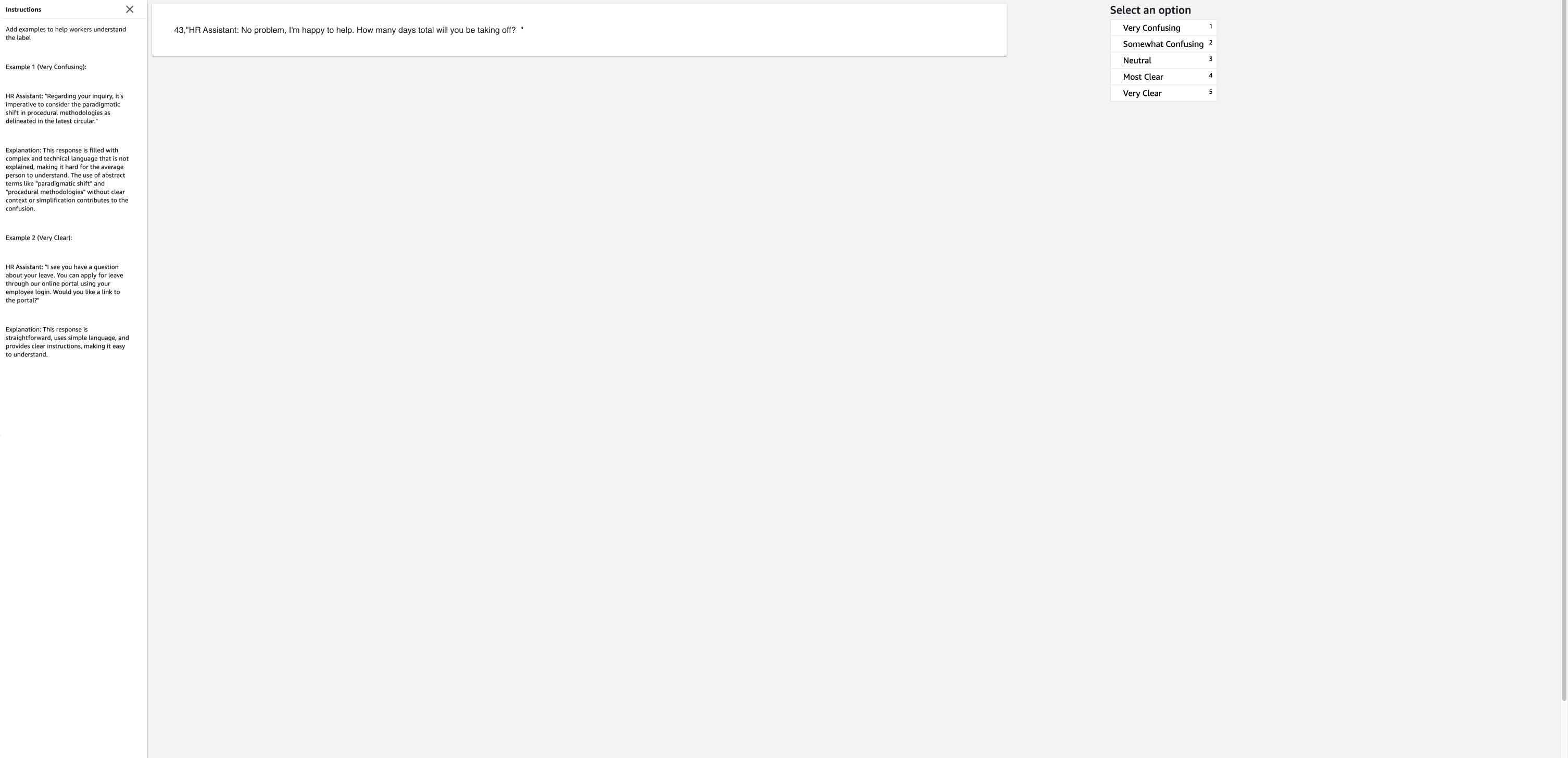}
\centering

\caption{MTurk human instructions to understand if the HR question is clear}
\end{figure*}

\begin{figure*}
\includegraphics[width=16cm]{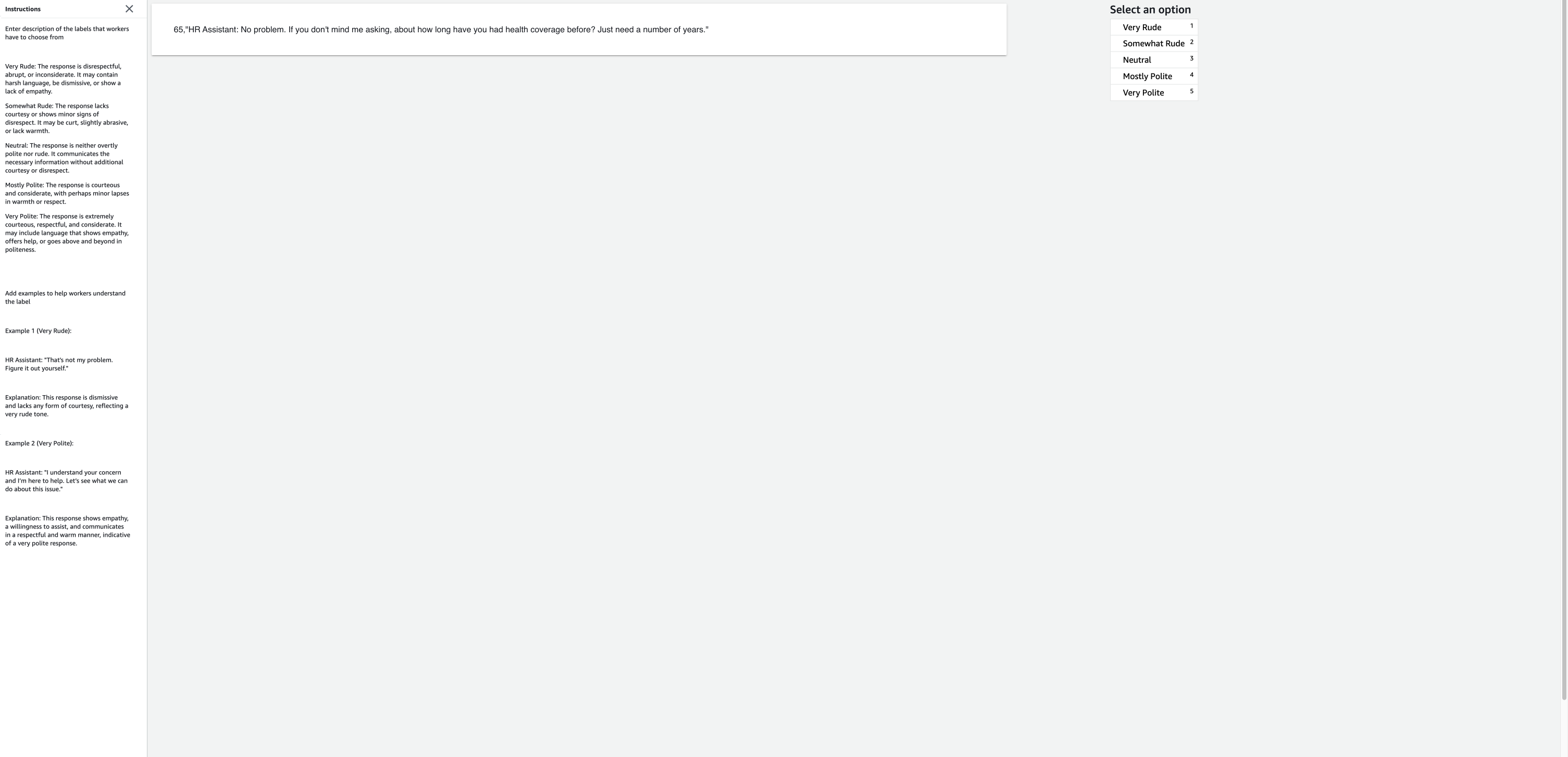}
\centering

\caption{MTurk human instructions to understand if the HR question is polite}
\end{figure*}

\begin{figure*}
\includegraphics[width=16cm]{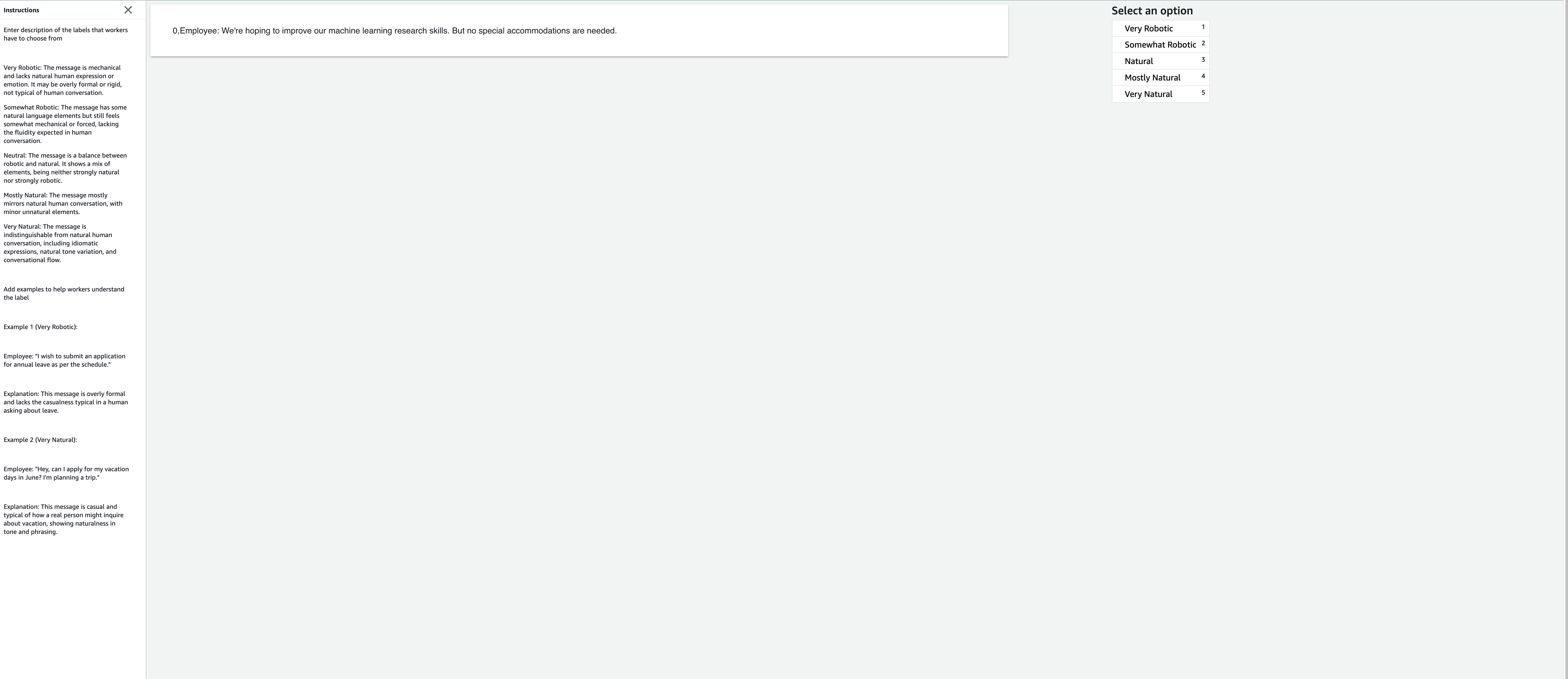}
\centering

\caption{MTurk human instructions to understand if the employee answer is natural}
\end{figure*}

\end{document}